\documentclass[journal,twoside]{IEEEtran}

\usepackage[ruled, linesnumbered, vlined, commentsnumbered]{algorithm2e}
\usepackage{amsfonts,amssymb,amsmath,amsthm,amsopn,mathrsfs,mathtools,wasysym}	
\usepackage{hhline,booktabs,colortbl,multirow,tabularx,diagbox,threeparttable} 
\usepackage[caption=false,font=normalsize,labelfont=sf,textfont=sf]{subfig}
\usepackage{graphicx} 
\usepackage{color,xcolor} 
\usepackage{arydshln}
\usepackage{enumerate}
\usepackage{authblk}
\usepackage{footnote}
\usepackage{hyperref}
\usepackage{prettyref}
\usepackage{cite}
\usepackage{setspace}
\usepackage{tcolorbox}
\usepackage{academicons}
\usepackage{xcolor}
\usepackage{orcidlink}
\usepackage{tabularx}
\usepackage{booktabs}
\usepackage{array}
\usepackage{bm}
\usepackage{csquotes}
\usepackage{mathtools}
\usepackage{blkarray, bigstrut} %
\usepackage{amsmath}

\definecolor{mycyan}{gray}{.7}

\newcommand{\pref}{\prettyref}

\newrefformat{fig}{Fig.~\ref{#1}}
\newrefformat{tab}{Table~\ref{#1}}
\newrefformat{sec}{Section~\ref{#1}}
\newrefformat{alg}{Algorithm~\ref{#1}}
\newrefformat{property}{Property~\ref{#1}}
\newrefformat{theorem}{Theorem~\ref{#1}}
\newrefformat{definition}{Definition~\ref{#1}}
\newrefformat{corollary}{Corollary~\ref{#1}}
\newrefformat{lemma}{Lemma~\ref{#1}}
\newrefformat{conj}{Conjecture~\ref{#1}}
\newrefformat{def}{Definition~\ref{#1}}
\newrefformat{eq}{equation~(\ref{#1})}
\newrefformat{app}{Appendix~\ref{#1}}

\newcommand{\orcid}[1]{\href{orcid.org/#1}{\textcolor[HTML]{A6CE39}{\aiOrcid}}}

\begin{document}

\title{Towards the Inferrence of Structural Similarity of Combinatorial Landscapes}

\author{Mingyu Huang$^{\orcidlink{0000-0003-2829-8673}}$ and
        Ke Li$^{\orcidlink{0000-0001-7200-4244}}$,~\IEEEmembership{Senior Member,~IEEE}
    \thanks{M. Huang is with James Watt School of Enginerring, University of Glasgow, G12 8QQ, Glasgow (e-mail: 2510648h@student.gla.ac.uk).}
    \thanks{K. Li is with Department of Computer Science, University of Exeter, Exeter, EX4 4QF, UK (e-mail: k.li@exeter.ac.uk).}
    }

\markboth{\uppercase{ieee transactions on evolutionary computation},~Vol.~x, No.~x, xxxx}{Huang and Li: On the Structural Similarity of Combinatorial Landscapes: A Graph Data Mining Pespective}

\maketitle

\begin{abstract} 
    One of the most common problem-solving heuristics is by analogy. For a given problem, a solver can be viewed as a strategic walk on its fitness landscape. Thus if a solver works for one problem instance, we expect it will also be effective for other instances whose fitness landscapes essentially share structural similarities with each other. However, due to the black-box nature of combinatorial optimization, it is far from trivial to infer such similarity in real-world scenarios. To bridge this gap, by using local optima network as a proxy of fitness landscapes, this paper proposed to leverage graph data mining techniques to conduct qualitative and quantitative analyses to explore the latent topological structural information embedded in those landscapes. By conducting large-scale empirical experiments on three classic combinatorial optimization problems, we gain concrete evidence to support the existence of structural similarity between landscapes of the same classes within neighboring dimensions. We also interrogated the relationship between landscapes of different problem classes.
\end{abstract}


\IEEEpeerreviewmaketitle

\section{Introduction}
\label{sec:introduction}

\IEEEPARstart{B}{lack-box} optimization problems (BBOPs), originated from black-box concepts in cybernetics, aims at finding a solution $\mathbf{x} \in \mathbb{R}^n$ that maximizes or minimizes an objective value $f(\mathbf{x})$ without any explicit knowledge of the objective function $f: \mathbb{R}^n \to \mathbb{R}$. Such problems are often encounterd in real-world engineering. For example, modern software systems are often highly configurable to allow users to tailor them to adapt to a variety of application scenarios. Tunning specific values of each configuration option in order to achieve the best functional properties is of vital importance in practice but often non-trivial. One of the complications can be attributed to the black-box nature of a configurable software system where the performance of a configuration is unknown before observing the response of the underlying system.

Although many successful BBOP solvers have been developed in different disciplines over the years, the \enquote{No-Free-Lunch} theorem~\cite{DolpertM97} points out that there is no single algorithm that could always yield superior performance than others for all optimization problems. Therefore, in-depth understanding of the characteristics of the underlying BBOP is essential to facilitate the design, selection and configuration of appropriate solvers for a given task~\cite{Malan21, ZouCLCJZ22, MalanE13, MunozSKH15, Kotthoff14, Smith-MilesL12, Smith-Miles08, QasemP10, Prugel-BennettT12}. However, since there does not exist any analytical form or authentic solution for a BBOP, it is hard to interpret and investigate except for an observation of input-output response through experimentation. 

The fitness landscape metaphor~\cite{Wright1932} has been extensively used for analyzing and understanding the characteristics of BBOPs in the meta-heuristic community given its \textit{structural characteristics} are able to reflect important properties of the underlying problem~\cite{Malan21, ZouCLCJZ22, MalanE13, MunozSKH15, Kotthoff14, Smith-MilesL12, Smith-Miles08}. In particular, there have been abundant evidence suggesting that certain patterns or sub-structures are shared among different instances of a class of problem~\cite{QasemP10, Prugel-BennettT12, Tayarani-NP16, Tayarani-Najaran15}, or, even across different classess of problems~\cite{TayaraniP14}.

In practice, one of the most common problem-solving approaches is by analogy, as it is particularly helpful for the understanding of abstract and sophiscated systems~\cite{RichlandS15}. In the context of BBOP, the basic assumption is that if a BBOP solver is effective for one problem, we could then expect its effectiveness for solving other problems whose fitness landscapes essentially share \textit{structural similarity} with each other. In other words, the strategic walk on the fitness landscape induced by the solver may be extended to landscapes of similar problems that are anticipated to share certain patterns or sub-structures but vary in different sizes or volumes. Thereby, when encountering a novel BBOP instance $\mathcal{P}_0$, it would be possible to study it by following an analogical reasoning process~\cite{Gentner83, ThagardHNG90}.

The inferrence of such similarity would thus not only enable us to transfer our knowledge on a well-studied problem to solve previously unseen instances, but also deepen our understanding of the relationships between different problem instances, and thereby brings a fundamental new pespective for studying BBOPs. However, given the extremely high-dimensional and abstract nature of BBOP landscapes, it is far from trivial to explicitly illustrate or quantitatively measure such similarity. Despite many efforts, there is a lack of dedicated work regard this topic due to the missing of effective analytical methods.

Over the last three decades and beyond, considerable efforts have been devoted to developing statistical measures to characterize fitness landscape as reviewed in~\cite{ZouCLCJZ22, Malan21, MalanE13}. Prominent example of these measures include autocorrelation~\cite{Weinberger1990} for characterizing landscape ruggedness and fitness distance correlation~\cite{JonesF95, MerzF00} to measure problem difficulty. 
Although informative, most of these features could only capture certain aspects of landscape property. Study in cogonitive science pointed out that it is often the case where high-order relationships play a much more important role in determing the similarity between two instances. For example, whilst battery and reservoir are superficially different things in many aspects (e.g., color, size shape and substance), at a higher level, their similary functions of storing and releasing energy makes it reasonble to consider them as analogous. It is thereby not sufficient to base our structural similarity inferrence only on landscape metrics.

In the more recent decade, local optima networks (LONs) have emerged as a powerful tool for modeling and analyzing fitnesss landscapes. Specifically, the vertices of a LON are local optima and the edges indicate certain search dynamics of a meta-heuristic algorithm. Since LONs are able to capture various characteristics of the underlying landscape (e.g., the number of local optima, their distribution and connectivity pattern), they are powerful tools for fitness landscape analysis. In particular, the graph nature of LON makes it highly suitable for studying the structural characteristics of fitness landscapes. For instance, graph layout and node embedding techniques would enable us to obtain intuitive visualizations of the typological structures of fitness landscapes and draw comparison between them~\cite{HuangL23}. In addition, graph representation learning techniques~\cite{ZhangYZZ20} are able to generate low-dimensional representations of fitness landscapes that are able to preserve high-level structural information, which have shown efficacy in the determination of structural similarity between different fitness landscapess~\cite{HuangL23}. 


Build on our recent work which conducted a case study on inferring the structural similarity of fitness landscapes across different dimensions of number partitioning problems (NPPs), in this study, we seek to gain further insights on a wider scope of landscapes with more comprehensive analysis. Specifically, we consider four types of classic BBOPs, inlcuding NPP, maximum satisfiability (Max-Sat), the traveling salesman problem (TSP), and the knapsack problem (KP). We seek to study the structural similarity of different problem instances at different levels, which lead us to our first three research questions (RQs). To start with, we also first consider problem instances of different dimensions. 

\begin{itemize}
    \item \textbf{\underline{RQ1}}: \textit{For each class of problems, can we investigate any potential structural similarity in fitness landscapes across instances of \underline{different dimensions?}}
    \item \textbf{\underline{RQ2}}: \textit{For each class of problems, can we investigate any potential structural similarity in fitness landscapes across instances of \underline{different sub-classes?}}
    \item \textbf{\underline{RQ3}}: \textit{Can we investigate any potential structural similarity in fitness landscapes across instances that belong to \underline{different classes} of problems?}
\end{itemize}




The empirical results with regard to the first three RQs is highly inspiring, as we are able to disclose the similarities and differences between a wide range of instances by both qualitative and quantitative analysis. We then evalute the efficacy of our measured similarity by running heuristic algorithms on each problem instance. The rationale behind this particular analysis is based on the hypothesis that the level of similarity between two landscapes should be correlated with the performance difference of meta-heuristics on the corresponding problems, which leads to our final RQ: 

\begin{itemize}
    \item \textbf{\underline{RQ4}}: \textit{How effective is the measured similarity in explaining the performance difference of meta-heuristics between different problem instances?}
\end{itemize}


The experimental results show that the $Sim$ value could has a good correlation with performance differences of testing algorithms.

Details of our empirical experiments and responses to these RQs will be elaborated in later sections step by step. Drawing on these answers, this study, for the first time, provides new insights help to further advance this field of research. In a nutshell, the main contributions of this paper could be summarized as follows:

\begin{itemize}
	\item To the best of our knowledge, this work represents the first attempt to infer structure similarity between fitness landscapes of different COPs.
	\item We proposed a new fitness landscape visualization technique based on LON by leveraging node embedding and dimensionality reduction method.
    \item We developed a quantitative measure of structural similarity between different combinatorial landscapes using graph embedding technique and correlation metrics.
    \item We developed a \texttt{Python} package for performing graph-based landscape analysis. 
\end{itemize}

In the rest of this paper, \pref{sec:preliminaries} provides some preliminary knowledge related to this work. \pref{sec:method} delineates our empirical methodologies for problem instance generation and LON construction. Extensive empirical results are presented and analyzed in~\pref{sec:results}. In the end, \pref{sec:conclusions} concludes this paper and threads some light on future directions.

\section{Preliminaries and Related Work}
\label{sec:previous_literature}

Over the last few decades, considerable efforts have been made to understand the properties of fitness landscapes of BBOPs. In this section, we first provide a brief review on the previous literatures on characterizing fitness landscapes. Then, we summarize exisiting methods for analyzing LON, which is most related to this paper. Finally, we review available techniques for visualizing fitness landscapes. 

\subsection{Problem Definitions}
\label{sec:problem_definitions}

\paragraph{Number partitioning problem (NPP)} 
This is a classic class of NP-complete problem~\cite{Korf98} that considers the task of deciding whether a given multiset $\mathcal{S}$ of positive integers can be divided into two disjoint subsets $\mathcal{S}^1$ and $\mathcal{S}^2$, i.e., $\mathcal{S}^1\cup \mathcal{S}^2=\mathcal{S}$, such that the sum of the numbers in $\mathcal{S}^1$ equals that in $\mathcal{S}^2$. A partition is called perfect if the discrepancy between the two subsets is $0$ when the sum of the original set is even, or $1$ when the sum is odd. Mathematically, the NPP considered in this paper is defined as follows:
\begin{equation}
    \min\ \left|\sum_{i=1}^{|\mathcal{S}^1|}\mathcal{S}^1_i-\sum_{i=1}^{|\mathcal{S}^2|}\mathcal{S}^2_i\right|.
    \label{eq:npp}
\end{equation}
In this paper, we use NPP-$n$ to denote the $n$-dimensional NPP instance whose cardinality is $n=|\mathcal{S}|$. For each NPP instance, the $n$ items on which partitioning is performed are randomly drawn within the range $[0,2^{n\cdot{k}}]$, where $k$ is a parameter that controls the hardness of the underlying NPP and it is set as $0.7$ in this study. 

\paragraph{Maximum Satisfiability Problem (Max-Sat)} 
The MAX-SAT problem is closely related to the satisfiability decision problem colloquially known as SAT. This problem involves a set of boolean variables $\mathbf{x}=(x_1,\dots,x_n)$ and a set of disjunctive clauses consisting of a subset of literals (a literal is either a variable or its negation). For example, a clause might be $x_1 \vee \neg x_5 \vee x_{10}$. Each clause can be considered an additional constraint that must be satisfied. In SAT the question is, "Does there exist an assignment of the variables which satisfies all the clauses?". A special variant of SAT is $k$-SAT which consists of clauses containing exactly $k$ literals. 

MAX-SAT is the generalization SAT to problems which are not fully satisfiable. It asks the question whether there exists an assignment of the variables which satisfies all but $T$ clauses. MAX-$k$-SAT is NP-hard for $k\geq2$ (thus MAX-$2$-SAT is NP-hard even though $2$-SAT is not). Throughout this paper we will restrict our attention to MAX-$3$-SAT. which is the best studied class of MAX-SAT problems. Assuming there are $n$ clauses and denoting them by $c_i(\mathbf{x})$, the objective is to:
\begin{equation}
    \max\ \sum_{i=1}^{n}\big\| g_i(\mathbf{x}) \,\, \text{is satisfied} \big\|.
    \label{eq:max_sat}
\end{equation}
where $\big\| g_i(\mathbf{X}) \,\, \text{is satisfied} \big\|$ is an indicator function equal to $1$ if clause $i$ is satisfied and $0$ otherwise.

Our focus will be on randomly generated instances, where each clause consist of $k$ ($3$ in this study) randomly chosen variables which are negated with probability of a half. We require each variable in a clause to be different and all clauses to be unique. We denote the number of variables by $n$, the number of clauses by $m$, and the ratio of clauses to variables by $\alpha$. Similar to the control variable $k$ in NPP, $\alpha$ here is also known to have significant impact on problem difficulty. We choose $\alpha=8$ in this study as in~\cite{QasemP10}.

\paragraph{Knapsack Problem (KP)}
In traditional KPs, i.e., bounded KP (BKP), a knapsack with capacity $C$ and $n$ types of objects are given. Each type has its profit, weight, and bound. Let $p_i$, $w_i$, and $b_i$ be the profit, weight, and bound of the $i$-th object, a solution of BKP is represented by a vector of length $n$, i.e., $\mathbf{x}=(x_1,\dots,x_n)$, where $x \in \{0,1,\dots,b_i\}$ is the number of objects taken from the $i$-th object. Then, the problem is mathematically modeled as
\begin{equation}
    \begin{aligned}
    &\max\ \sum_{i=1}^{n}x_i p_i \\
    &\text{s.t.}  \begin{cases}
        \forall i \in \{1,2,\dots,n\},x_i \in \{0,1,\dots,b_i\} \\
        \sum_{i=1}^{n} x_iw_i \leq C 
      \end{cases}
    \end{aligned}
    \label{eq:kp}
\end{equation}
In this study, we focus on $0$-$1$KP, which is a special case of BKP with $b_i=1, i=1,\dots,n$, and has been widely studied. We generate random instances of $0$-$1$KP based on the strategies proposed in~\cite{Pisinger05}. In particular, we based our studies on the class of weakly correlated $0$-$1$KP (WKP), where the profit $p_i$ of each object has a weak correlation with its weight $w_j$. Let the weight $w_j$ of the $i$-th item drawn randomly from range $[1,R]$, the profit $p_i$ will then be chosen randomly from range $[w_i - R/10, w_i + R/10]$ such that $p_i \geq 1$. The capacity $C$ is determined as $0.55W$, where $W = \sum_{i=1}^{n}b_iw_i$ (note that $b_i=1$ in $0$-$1$KP). $R$ is set to $1,000$ as in~\cite{WangZ21}.

\subsection{Fitness Landscape Analysis}

The original idea of fitness landscape dates back to 1932 when Wright pioneered this concept in evolutionary biology~\cite{Wright1932}. Since then, fitness landscapes have been extensively studied in both physics and optimization communities. In the field of evolutionary computation, a large set of measures have been developed to characterize features of fitness landscapes since the 1990s. 
For instance, Jones and Forrest~\cite{JonesF95} proposed a measure of landscape \textit{neutrality} based on the correlation between distances from a set of points to the global optimum and their corresponding fitness values. Other measures of neutrality include neutral walks~\cite{ReidysS2001} and neutral network analysis~\cite{VanneschiPC06}. 
In addition, adaptive walks~\cite{KauffmanL1987}, autocorrelation measures~\cite{Weinberger1990}, correlation length~\cite{ManderickWS1991} as well as entropic measures~\cite{MalanE09} are developed to measure the \textit{ruggedness} of a fitness landscape. 
Lunacek and Whitley~\cite{LunacekW06} introduced \textit{dispersion} features, which compare pairwise distances of all points in the initial design with the pairwise distances of the best points in the initial design. 
\textit{Epistasis} metrics, such as epistasis variance~\cite{Davidor90}, epistasis correlation~\cite{RochetVSK97}, and bitwise epistasis~\cite{FonluptRP98}, have been used to determine the interaction between variables.
In other works, Morgan and Gallagher analyzed optimization problems by means of \textit{length scale} features~\cite{MorganG17},  Shirakawa and Nagao introduced the \textit{bag of local landscape features}~\cite{ShirakawaN16}. Munoz \textit{et al.} measured the change in the objective values of neihgboring points to determine the landscape's \textit{information content}~\cite{MunozKH12,MunozKH15}.
Many of the above features were synthesized into the concept of \textit{exploratory landscape analysis} (ELA)~\cite{MersmannBTPWR11} and the \texttt{R}-package \texttt{flacco}~\cite{Kerschke17}, which have been widely adopted by many recent works~\cite{Tanabe22,RehbachZFRB22,KerschkeT19,SchneiderSPBTK22}. 

\subsection{Local Optima Network Analysis}

LONs are rooted in the study of energy landscapes in chemical physics~\cite{Stillinger95} and are first introduced to the analysis of Kauffman's $NK$-landscapes in a series of work around 2010~\cite{TomassiniVO12,OchoaTVD08,VerelOT08,VerelOT11}. In successive studies, LONs are then applied to analyze the landscapes of various COPs, including quadratic assignment problem (QAP)~\cite{DaolioVOT10}, traveling salesman problem~\cite{OchoaVWB15}, number partitioning problem (NPP)~\cite{OchoaVDT17}, etc. More recently, applications of LONs have been seen in assisting the understanding of real-world configurable system, such as neural architecture search (NAS)~\cite{PotgieterCB22}, automated machine learning (AutoML)~\cite{TeixeiraP22, HuangL23b} and computational protein design~\cite{SimonciniBSV18}. 

There exist various types of definitions of an edge in LONs, e.g., the transition probabilities between basins~\cite{OchoaTVD08}, the escape probability between local optima through perturbations~\cite{VerelDOT11} and the transition between local optima via crossover~\cite{OchoaCTW15}. LONs could be trimmed by only keeping the non-deteriorating moves, which result in the so-called \textit{monotonic local optima network} (M-LON). By compressing the plateaus (i.e., sets of local optima wtih equal fitness values) into single nodes, M-LON could be further simplified into \textit{compressed monotonic local optima network} (CM-LON). 

Existing approaches for analyzing LONs are mainly dependent on metrics developed for complex networks~\cite{BoccalettiLMCH06}, e.g., degree, clustering coefficient, centrality measures, path length as well as assortativity, etc.  

\subsection{Fitness Landscape Visualization}

Due to the high-dimensional nature of combinatorial landscapes, it is challenging to generate informative $2$-D or $3$-D visualizations of their topological structures. Over the years, limited work has been done regard this matter. 

In recent years, state-of-the-art algorithms for dimensionality reduction techniques such as t-distributed stochastic neighbor embedding (t-SNE)~\cite{MaatenH08} and uniform manifold approximation and projection (UMAP)~\cite{McInnesH18} have shown their capability for visualizing high-dimensional landscapes. Michalak~\cite{Michalak19} proposed a method for combinatorial landscape visualization by combining t-SNE and vacuum embedding. This is similar to the work by Shires and Pickard~\cite{ShiresP21}, in which a technique based on UMAP is used to generate low-dimensional plots of energy landscapes.

Due to the graph nature of LONs, it is not surprising that they could enable insightful visualization of fitness landscapes. The most direct way for presenting LONs is directly draw the nodes and their connections~\cite{DaolioTVO11}, where different colors and sizes could be assigned to nodes/edges to indicate interested attributes. An obvious drawback of this method is that it is only applicable to problems of very low dimensions, since with the increase in the number of local optima, the output will quickly become too messy for identifying useful information. By applying certain layout strategies, this could be alleviated to some extend~\cite{OchoaV16,VeerapenO18}. In addition, $3$-D visualizations~\cite{OchoaVDT17,VeerapenDO17} \dots.


\section{Propose Analysis Framework}
\label{sec:method}

\begin{figure*}[t!]
    \centering
    \includegraphics[width=.8\linewidth]{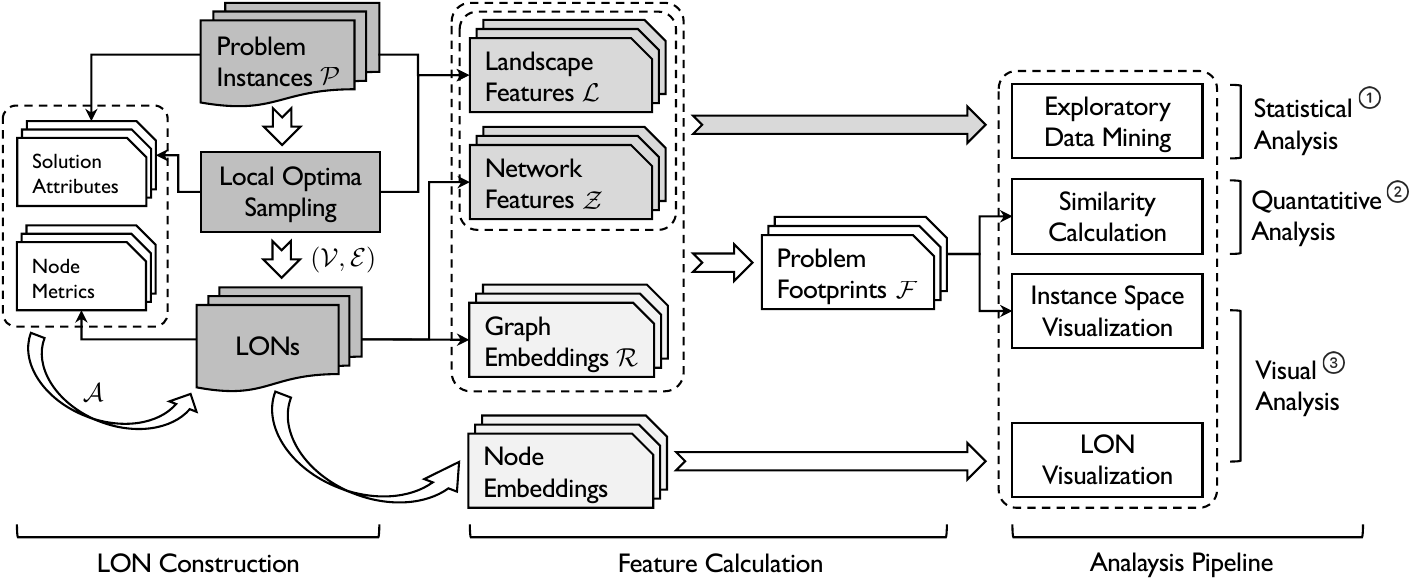}
    \caption{Framework.}
    \label{fig:framwork}
\end{figure*}

\subsection{Framework}

Although many efforts have been devoted to the application of LONs on FLA, there is no dedicated software package for end-to-end LON-based landscape analysis. Also, to address the RQs posed in~\pref{sec:introduction}, a set of different analysis approaches is demanded to a obtain comprehensive understanding of the topological structures of LONs, and preferably, from  different aspects. In addition, as the first step towards the inferrence of structural similarity between different fitness landscapes, it would be desirable to formulize the analytical pipeline and synthesize it into an open-source pacakge, and hence allow this to be accessible by a wider community of researchers. Bearing these considerations in mind, this article proposes an analysis framework for exploring and comparing the structural characteristics of fitness landscapes of COPs along with a highly integrated \texttt{Python} package, \texttt{GBFLA}. 

The high-level structure of the proposed framework is shown in~\pref{fig:framwork}, from which we could see that the whole analysis starts out with the choosing of a set of problem instances $\{\mathcal{P}_1, \dots, \mathcal{P}_n\}$. Thereafter, we apply local search strategies to draw representative samples of local optima as well as the transitions among them from the search space $\mathcal{X}$. The obtained data is then used to construct LONs for fitness landscape of each problem instance respectively. After that, we use the developed LONs as the driver to conduct exploratory analysis on the corresponding fitness landscapes to investigate and compare their structural properties. As shown in~\pref{fig:framwork}, the proposed analysis pipeline consists of three types of approaches, which are supported by different types of calculated features, including a \textit{statistical} analysis based on network and landscape features, a \textit{visual} analysis of LONs, as well as a \textit{quantitative} determination of the similarity between LONs of different landscapes. Together these analysis are able to provide a thorough understanding and comparision of the underlying structure of LONs and thus the associated fitness landscapes. We will delineate the details for each analysis method as well as the corresponding features in later subsections. 

All the modules presented in~\pref{fig:framwork} are implemented in the accompanied package \texttt{GBFLA}, and at its core lies the \texttt{LON} object, which stores the information contained in a LON such as nodes, edges and node attributes. In particular, in \texttt{GBFLA}, LON data is accessible as either \texttt{pandas} dataframe (\texttt{LON.data}) or \texttt{NetworkX} graph (\texttt{LON.graph}) object, and it is thus naturally suitable for data mining and manipulation. We believe this tool as well as the proposed graph-based analysis framework could not only set up a standard routine for the inferrence of strucutral similarity between combinatorial landscapes, but also benefit researchers who are interested in a wider range of tasks. For instance, the proposed visualization method could provide a fundamental new illustration of the distribution and connectivity pattern of local optima in a fitness landscape, and thus assist the design of problem solvers. In addition, LON embeddings generated using network representation learning methods could constitute feature vectors for tasks like automated algorithm selection and performance prediction. 

\begin{algorithm}[t]
    \caption{ILS for sampling local optima}
    \label{alg:ILS}
    
    \KwIn{Search space $\mathcal{X}$, fitness function $f$}
    \KwOut{$\mathcal{V}$, $\mathcal{E}$}
    $\mathcal{V}\leftarrow\emptyset$, $\mathcal{E}\leftarrow\emptyset$\;
    Generate an initial candidate $\mathbf{x}\in\mathcal{X}$ by random sampling\;
    $\mathbf{x}^\ell\leftarrow\mathtt{localSearch}(\mathbf{x})$\;
    $\mathcal{V}\leftarrow\mathcal{V}\cup\{\mathbf{x}^\ell\}$\;
    $i\leftarrow 0$\;
    \While{$i\leq K$}{
        $\mathbf{x}'\leftarrow\mathtt{perurbation}(\mathbf{x}^\ell)$\;
        $\mathbf{x}^{\ell'}\leftarrow\mathtt{localSearch}(\mathbf{x}')$\;
        \If{$f(\mathbf{x}^{\ell'})\leq f(\mathbf{x}^{\ell})$}{
            $f(\mathbf{x}^\ell)\leftarrow f(\mathbf{x}^{\ell'})$\;
            $\mathcal{V}\leftarrow\mathcal{V}\cup\{\mathbf{x}^\ell\}$\;
            Construct an edge between $\langle\mathbf{x}^\ell,\mathbf{x}^{\ell'}\rangle$ and add it to $\mathcal{E}$\;
            $i\leftarrow 0$;
        }
        $i\leftarrow i+1$;
    }
    \Return{$\mathcal{V}$, $\mathcal{E}$}
\end{algorithm}

\subsection{Local Optima Sampling}

As shown in Fig. X, for a given set of problem instances, to construct LONs of their fitness landscapes, we first need a proper sampling process to extract representative samples of local optima (as node set $\mathcal{V}$) and determine their connectivity patterns (as edge set $\mathcal{E}$). In our implementation, we apply the widely used ILS, a powerful local search meta-heuristic~\cite{ILS,ILS_Powerful}, to serve this purpose~\cite{ILS_Standard1,ILS_Standard2}
\footnote{There are alternative methods for sampling local optima from combinatorial search spaces, e.g., hybrid genetic algorithm (GA) sampling, snowball sampling, and interested readers are refered to~\cite{VeerapenO18,ThomsonOV19}}.
The persudo code for ILS is shown in~\pref{alg:ILS}. Generally, the algorithm starts by randomly initializing a $n$-dimensional vector $\mathbf{x}$ as the initial solution, and then local search is applied to $\mathbf{x}$ until a local optimum $\mathbf{x}^{\ell_{0}}$ is found. Afterwards, a \textit{perturbation} is applied on $\mathbf{x}^{\ell_{0}}$ to escape to a new solution $\mathbf{x'}$, and the algorithm then starts climbing from $\mathbf{x'}$ until a new local optima $\mathbf{x}^{\ell_{1}}$ is reached. An edge traced from $\mathbf{x}^{\ell_{0}}$ to $\mathbf{x}^{\ell_{1}}$ will then be recorded to the edge set $\mathcal{E}$, and the corresponding local optima are added to the node set $\mathcal{V}$. This perturbation process is repeated until a specified termination condition is reached (e.g., \texttt{max\_nimpr} = $100$ non-improvement perturbations). Thereafter, another ILS will be restarted from a new randomly initialized solution. To sample sufficient local optima for each problem instance, such random initializations will be performed many times (e.g., \texttt{n\_iter} = $1000$). During the whole sampling process, the visiting frequency of each transition is recorded as the weight for each edge (e.g., the transition $\mathbf{x}^{\ell_{3}} \to \mathbf{x}^{\ell_{9}}$ may have occured $48$ times during a search process). Other interested information such as the number of hill-climb steps taken to reach a local optimum, are also recorded during ILS, which could later serve as node attributes for LON later. 
  
\subsection{LON Construction}
\label{sec:method_lon}

\begin{table}
    \footnotesize	
    \centering
    
    \caption{Low-level features}
    \resizebox{.48\textwidth}{!}{
    \begin{tabularx}{\linewidth}{
        p{\dimexpr.13\linewidth-2\tabcolsep}
        p{\dimexpr.87\linewidth-2\tabcolsep}
        }
        \hline
        \addlinespace[2pt]
        Symbol & Description \\
      \addlinespace[2pt]
      \hline
      \addlinespace[3pt]
      $b_{\text{size}} \dag$     & Size of basin of attraction of a \textit{lo}, approximated using the sampling strategy proposed in~\cite{OchoaVWB15}. \\
      \addlinespace[1pt]
      $n_{\text{climb}} \dag$    & Number of hill-climb steps taken to reach a \textit{lo}, recorded in ILS. \\
      \addlinespace[1pt]
      $n_{\text{pert}} \dag$     & Number of perturbations taken to find an improving move from a \textit{lo}, recorded during ILS.  \\
      \addlinespace[1pt]
      $freq \dag$                & Frequency of visits to a \textit{lo} during ILS. \\
      \addlinespace[2pt]
      \hline\hline
      \addlinespace[2pt]
      $deg \ast$                & Degree of a node, i.e., number of neighbours of a node.  \\
      $deg_{\text{in}} \ast$    & Incoming degree, i.e., number of \textit{lo}s that could transit to the target one via perturbation followed by hill-climbing. \\
      $deg_{\text{out}} \ast$   & Outgoing degree, i.e., number of \textit{lo}s that could be reached from a source \textit{lo} via perturbation followed by hill-climbing. \\
      $c_{\text{betw}} \ast$     & \textit{Betweeness centrality}, which is a measure of how often a node lies on the shortest path between other nodes in a network. Nodes with high betweenness centrality can have a large impact on the flow of information through the network. \\
      \addlinespace[1pt]
      $c_{\text{ev}} \ast$       & \textit{Eigenvector centrality}, this is a measure of a node's importance in a network based on the importance of the nodes it is connected to. Nodes with high eigenvector centrality are connected to other nodes that are also important in the network.\\
      \addlinespace[1pt]
      $c_{\text{close}} \ast$    & \textit{Closeness centrality}, a measure of a node's importance in a network based on the number and quality of incoming links to the node. Originally developed for ranking web pages in search engines, PageRank centrality can be applied to other types of networks as well.\\
      \addlinespace[1pt]
      $c_{\text{pg}} \ast$       & \textit{PageRank centrality}, this is a measure of a node's importance in a network based on the number and quality of incoming links to the node. Originally developed for ranking web pages in search engines, PageRank centrality can be applied to other types of networks as well. \\
      \addlinespace[1pt]
      $cc \ast$                  & \textit{Clustering coefficient}. It quantifies the degree to which nodes in a network tend to cluster together.  \\
      \addlinespace[1pt]
      $avg_{\text{deg}} \ast$    & Average degree of the neighbours of a \textit{lo}.  \\
      \addlinespace[1pt]
      $avg_{\text{fit}} \ast$    & Average fitness of the neighbours of a \textit{lo}. \\
      \addlinespace[1pt]
      $l_{\text{min}} \ast$      & Mininum length to accessible global optima.  \\
      \addlinespace[1pt]
      $l_{\text{avg}} \ast$      & Average length to accessible global optima.  \\
      \addlinespace[2pt]
      \hline
    \end{tabularx}
    }
    \begin{tablenotes}
        \scriptsize
        \item[] $\dag$ denotes landscape features associated with each \textit{lo}.
        \item[] $\ast$ denotes network metrics calculated for each node. 
    \end{tablenotes}
    \label{tab:low_level_features}%
\end{table}

After obtaining the node set $\mathcal{V}$ and edge set $\mathcal{E}$, we could use them to construct LON as a directed and weighted graph, where the edge direction represents transition direction and weight indicates visiting frequency of a local optimum during ILS. In addition, we incorporate a set of node-level features (see~\pref{sec:feature_set}) $\mathcal{A}$ into LON. Each LON object in \texttt{GBFLA} is accessible in both graph (\texttt{GBFLA.LON.graph}) and pandas dataframe form (\texttt{GBFLA.LON.data}), and thus is naturally convenient for data mining and manipulation. 

\subsection{Feature Set Construction}
\label{sec:feature_set}
We develop landscape/LON features at two levels, as shown in the central part of~\pref{fig:framwork}. Firstly, for a given LON, we calculate:
\begin{itemize}
    \item A series of landscape properties (e.g., size of basin of attraction), as shown in the first block in~\pref{tab:low_level_features}. 
    \item A collection of node-level metrics calculated from LONs, as shown in the second block in~\pref{tab:low_level_features}. 
    \item A feature vector of each node generated by node embedding technique. 
\end{itemize}
These together then constitute an attribute vector $\mathbf{p}$ for each node, and are refered to as \textit{low-level} or \textit{node-level} features. Notably, the feature representation of each node obtained using node embedding methods could preserve the intrinsic properties of the original network, i.e., nodes that are connected to each other are put close in the vector space. Therefore, such embeddings are essentially useful for generating low-dimensional visualizations of LON or assisting downstream tasks such as clustering and community detection. 

In addition to these low-level features, we also characterize each fitness landscape using a set of \textit{high-level}, or say, \textit{graph-level} features, which is denoted by a feature vector $\mathbf{z}$. More specifically, it consists of three components, including: 
\begin{itemize}
    \item A set of high-level landscape features, e.g., fitness distance correlation, as shown in the first block in~\pref{tab:high_level_features}. 
    \item A set of high-level LON features, e.g., network density, as shown in the second block in~\pref{tab:high_level_features}.
    \item Feature representation $\mathcal{R}$ of each LON generated by graph embedding technique. 
\end{itemize}
Since this feature vector $\mathbf{z}$ is able to constitute a comprehensive description of the properties of a given fitness landscape, we also denote it as the \textit{footprints} of the landscape. Notably, this is the first time that graph embeddings $\mathcal{R}$ extracted from LON is introduced as the feature set for characterizing fitness landscape, which could capture certain topoglical structure of LON.

\subsection{Statistical Analysis}

\begin{table}
    \small
    \centering
    
    \caption{High-level features}
    \resizebox{.48\textwidth}{!}{
    \begin{tabularx}{\linewidth}{
        p{\dimexpr.15\linewidth-2\tabcolsep}
        p{\dimexpr.85\linewidth-2\tabcolsep}
        }
        \hline
        \addlinespace[2pt]
        Symbol & Description \\
      \addlinespace[2pt]
      \hline
      \addlinespace[3pt]
      $B_{\text{size}} \dag$     & Average size of basin of attraction. \\
      $N_{\text{climb}} \dag$    & Average number of hill-climb steps taken to reach a \textit{lo}. \\
      $N_{\text{pert}} \dag$     & Average number of perturbations taken to find an improving move from a \textit{lo}.  \\
      $N_{\text{funnel}} \dag$   & Number of funnels in the landscape.  \\
      \addlinespace[2pt]
      \hline\hline
      \addlinespace[2pt]
      $N_{\text{node}} \ast$     & Number of nodes in LON.  \\
      $N_{\text{edge}} \ast$     & Number of edges in LON.  \\
      $dens \ast$                & Network density.  \\
      $cc_{\text{avg}} \ast$     & Average clustering coefficient. \\
      $ast_{\text{deg}} \ast$    & Degree assortativity coefficent. \\
      $ast_{\text{fit}} \ast$    & Fitness assortativity coefficent. \\
      $L_{\text{avg}} \ast$      & Mean average path lengths to global optima. \\
      $L_{\text{min}} \ast$      & Mean minimum path lengths to global optima. \\
      \addlinespace[2pt]
      \hline\hline
      \addlinespace[2pt]
      CDD$\star$                 & Cummulative degree distribution trajectory of a LON  \\
      RCC$\star$                 & Rich club coefficent trajectory of a LON   \\
      \addlinespace[2pt]
      \hline
    \end{tabularx}
    }
    \begin{tablenotes}
        \scriptsize
        \item[] $\dag$ denotes landscape features.
        \item[] $\ast$ denotes network metrics calculated for each LON. 
        \item[] $\star$ denotes trajectories calculated for each LON
    \end{tablenotes}
    \label{tab:high_level_features}%
\end{table}

As the first step of our proposed analytical pipeline, we conduct a statistial exploration on LONs and the corresponding fitness landscapes. In particular, we first calcualte two sets of features which are able to capture high-level properties of fitness landscapes, as described in~\pref{tab:high_level_features}. We plot the trajectories of these features with the increase in dimensionality for each problem respectively. This allow us to compare the high-level characteristics of different combinatorial landscapes. It is worthy to note that though we have based this analysis mainly on network metrics, many traditional ELA features (e.g., fitness distance correlation from Table X), which could be easily calculated using the \texttt{R} package \texttt{flacco}, could also be introduced into this pipeline to consitute a richer set of features dependent on the tasks at hand. 

To obtain more detailed insights into each fitness landscape, we inspect the correlation between node attributes introduced in~\pref{tab:low_level_features} and the fitness values of local optima. These could enable comparison of the intrinsic structural properties of each landscape.

\subsection{Visual Analysis}

Graphical representation of LONs could provide insight into the structure of the fitness landscape of a given optimization problem. However, as reviewed in~\pref{sec:previous_literature}, existing approaches toward this end usually suffer from poor scalability. In this paper, we apply a novel method for visualizing LONs, which is based on NRL and dimnensionality reduction techniques. Such graph visualization strategy has shown its effectiveness in other disciplines~\cite{GoyalF18}. A specified node embedding techique would first generate a low-dimensional feature representation $\mathbf{Z}_i$ for each node in a LON. By doing so, the network will be transformed into a $|\mathcal{V}|\times|\mathbf{Z}_i|$ matrix. We then apply dimensionaliy reduction technique (e.g., UMAP~\cite{McInnesH18}) to further compress this embedding into $2$-dimension to enable visualization of the LON nodes in a bivariate space. 

Alternatively, to more directly visualize the \lq structural closeness\rq of LONs of a set of different problem instances, we could plot different problem instances in a $2$-dimensional \textit{instance space} based on a set of landscape or LON features, as inspired by the concept of \textit{instance space analysis}~\cite{SmithMilesT12,IclanzanDT14}. 
Instead of using hand-crafted features as in other works~\cite{SmithMilesT12,IclanzanDT14}, we adopt graph embedding techique to automatically extract feature vectors for each LON and thus constitute footprints for each landscape. These features are able to preserve the topological information of complex networks and are thus idea choices for studying structural similarity. 

\subsection{Quantitative Measurement of Structural Similarity}

We propose to use the Spearman correlation coefficient to measure the inter-correlation between the feature vectors generated using graph embedding technique. The resulted correlation value could play as a metric (denoted as $\mathrm{Sim}$) to quantitatively evaluate the structural similarity or \lq closenss \rq between different fitness landscapes.

\subsection{Performance Measures}

To answer \textbf{RQX}, a quantitative and meaningful performance measure of heuristic algorithm is essential. In this study, we base the performance assessment on algorithm \textit{runtime} studied in~\cite{HansenABT22}, which is more preferable compared to CPU or wall-clock time in terms of reproducibility and comparability across different problems. In particular, we adopted \textit{expected runtime} (ERT), which measures the expected number of function evaluations (FEs)~\cite{Tanabe22,AugerH05} needed to reach a target value $f_{\text{target}}$ and could be calculated by:
\begin{equation}
    ERT = \frac{\sum_{i=1}^{N_{\text{run}}}FE_i}{N_{\text{succ}}}
    \label{eq:ert}
\end{equation}
where $FE_i$ is the number of all FEs conducted in the $i$-th function instance until the algorithm terminates (after reaching $f_{\text{target}}$). $N_{\text{run}}$ is the nubmer of runs, $N_{\text{succ}}$ is the number of successful runs. We say that a run is successful if the algorithm reaches $f_{\text{target}}$. In this study, $f_{\text{target}}$ is set as the fitness value of top $1\%$ sampled local optima.

\section{Experimental Setup}
\label{sec:setup}

This section introduces the setup of our experiments, include the problem instances generated and the detaile configuration of the analysis pipeline proposed in~\pref{sec:method}. 

In order to answer the RQs posed in~\pref{sec:introduction}, we choose to study the three classic COPs, namely NPP, TSP and $0$-$1$KP introduced in~\pref{sec:problem_definitions}. For any given problem, we generate $30$ random instances with dimension varies within $n \in {\{10,13,15,17,20,23,25\}}$, which is adequate to illustrate patterns in fitness landscpaes across a range of dimensions. For each problem with a given $n$, we apply the ILS sampling process with \texttt{n\_iter} $=10000$ independent iterations, where for each run, the termination condition is \texttt{max\_nimpr} $=100$ non-improvement perturbations. For all the three problems, we adopt \textit{best-improvement} hill climbing as the local search strategy for ILS, and the perturbation operator is based on \textit{two-bit-flip} mutation. The results of all ILS iterations all combined to constitute the corresponding LON as introduced in~\pref{sec:method_lon}. In particular, we adopt an \textit{acceptance criterion} that only keeps non-deteriorating transitions between local optima, i.e., $\mathbf{x}^{\ell_{i}} \to \mathbf{x}^{\ell_{j}}$, where $\mathbf{x}^{\ell_{j}} \ge \mathbf{x}^{\ell_{i}}$ for maximization problems or $\mathbf{x}^{\ell_{j}} \le \mathbf{x}^{\ell_{i}}$ for minimization problems~\cite{ThomsonDO17}. This could serve as a filter for the ILS perturbation process and reduce the complexity of the LON, allowing us to focus our attention on the most critical transitions (i.e., improving moves). Finally, $7 \times 3 \times 30 = 630$ ($7$ dimensions $\times$ $3$ problems $\times$ $30$ random instances) LONs will be constructed. 

In \texttt{GBFLA}, various sub-components are configurable, e.g., correlation metrics, NRL methods, dimensionality reduction technique. Througout this paper, we use Spearman correlation as the measure of correlation. The reasoning behind this is that we observed that most relationship 


\section{Results and Analysis}
\label{sec:results}

\begin{figure*}[t!]
    \centering
    \includegraphics[width=\linewidth]{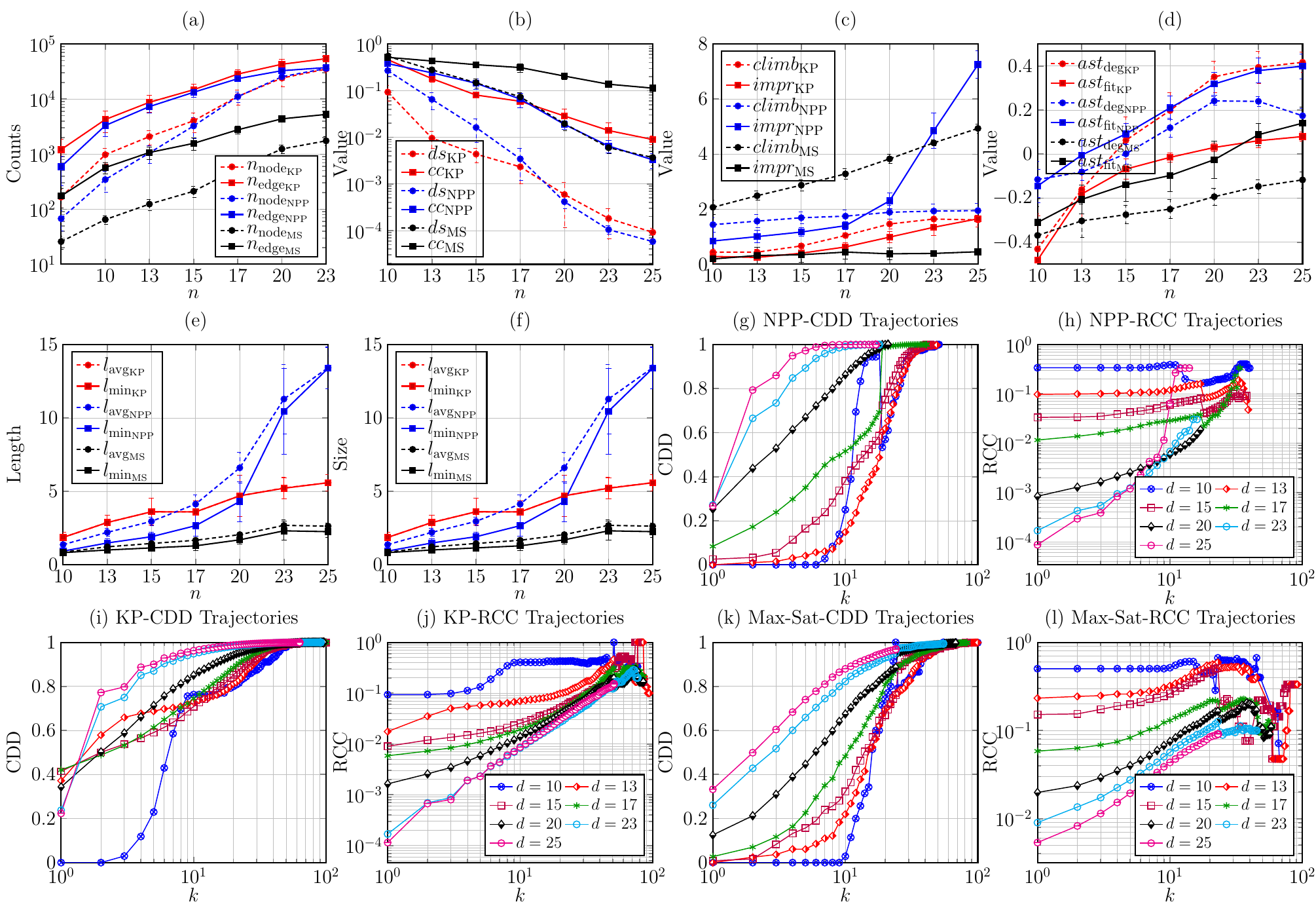}
    \caption{xxx}
    \label{fig:metrics}
\end{figure*}

In this section, we seek to address the three RQs raised in~\pref{sec:introduction} through a series of dedicated experiments. In particular, we present our empirical results and analysis on investigating the structural similarity of fitness landscapes with respect to three types of COPs using the LON-based FLA framework proposed in~\pref{sec:method}. We then verify the effectiveness of our proposed structural similarity measure by running heuristic algorithms on the studied problem instances and comparing their performance differences. 

\subsection{Structural Similarity Study for a Given Problem by Statistical Analysis}
\label{sec:rq1}

\subsubsection{Results for high-level statistial analysis}
\label{sec:results_rq1_1}

\pref{fig:metrics}(a) to \pref{fig:metrics}(f) plot the trajectories of the high-level features of LONs/fitness landscapes with the increase in dimensionality $n$ for different problems respectively. From the results, we could clearly see that for all three problems, most indicators experience a monotonic variation with the increase of $n$, which indicate that LONs for a given problem with similar dimensions tend to possess similar properties. More specifically:

\begin{itemize}
    \item As shown in~\pref{fig:metrics}(a), the number of both nodes and edges in LONs for all three problems increase at a logarithmic scale with the increase of problem dimensions. In particular, it could be observed that LONs of KP and NPP with similar dimensions are close in network size, while MaxSat tends to have much smaller LON than them at the same dimension.
    \item From~\pref{fig:metrics}(b), we could see that the density and average clustering coefficient of LONs of the three problems all decrease quickly as $n$ becomes higher. This is reasonable since in~\pref{fig:metrics}(a), the difference between number of nodes and edges of LONs only 
    \item \pref{fig:metrics}(c) shows the average number of hill-climbing steps and perturbations taken taken to reach a local optima / escape from the previous optimum. We could notice that generally, for all three problems, the hill-climb steps required to reach a local optimum tend to increase with dimensionality. Such trend is most perceivable for MaxSat, and is not obvious for NPP. As for perturbations, the number of perturbations needed to find an improving move increase dramatically with $n$. This is also true for KP, but with a much smaller slope, whereas for MS, it is hard to observe pattern regard this matter. 
    \item It could be observed from \pref{fig:metrics}(d) that the degree assortativity of LONs of NPP and KP undergo a monotonic change from negative values to positive ones. This is interesting, since it indicates that for low-dimensional problems, there are disassortative mixings in the network, i.e., local optimum with dissimilar degrees tend to connect with each other. As $n$ increases, this situation changed, and nodes with similar degree are likely to form connections. However, for MaxSat, the degree assortativity remains negative for all studied dimensions, yet the absolute values decreases as $n$ goes higher. As for fitness assortativity, the 
    \item \pref{fig:metrics}(e) depicts the average and minimum length to the accessible global optima(um) of each node. We could see that for all three problems, the path lengths to global otpima(um) increases with problem dimension. This is most notable for NPP. Although under a given dimension, LONs of KP tend generally have similar network sizes with NPP, the two trajectories for KP are substantially lower than those of NPP. Finally, the trajectories of MaxSat have even lower values, which is reasonable as LONs are much smaller in sizes compared to the other two problems. 
\end{itemize}

As for the CDD and RCC trajectories shown in~\pref{fig:metrics}(g) to~\pref{fig:metrics}(l), we have similar observations as above, i.e., a given problem tends to produce similar trajectories when the dimensionality is close to each other. 

\subsubsection{Results for low-level statistial analysis}
\label{sec:results_rq1_2}

\begin{figure}[t!]
    \centering
    \includegraphics[width=\linewidth]{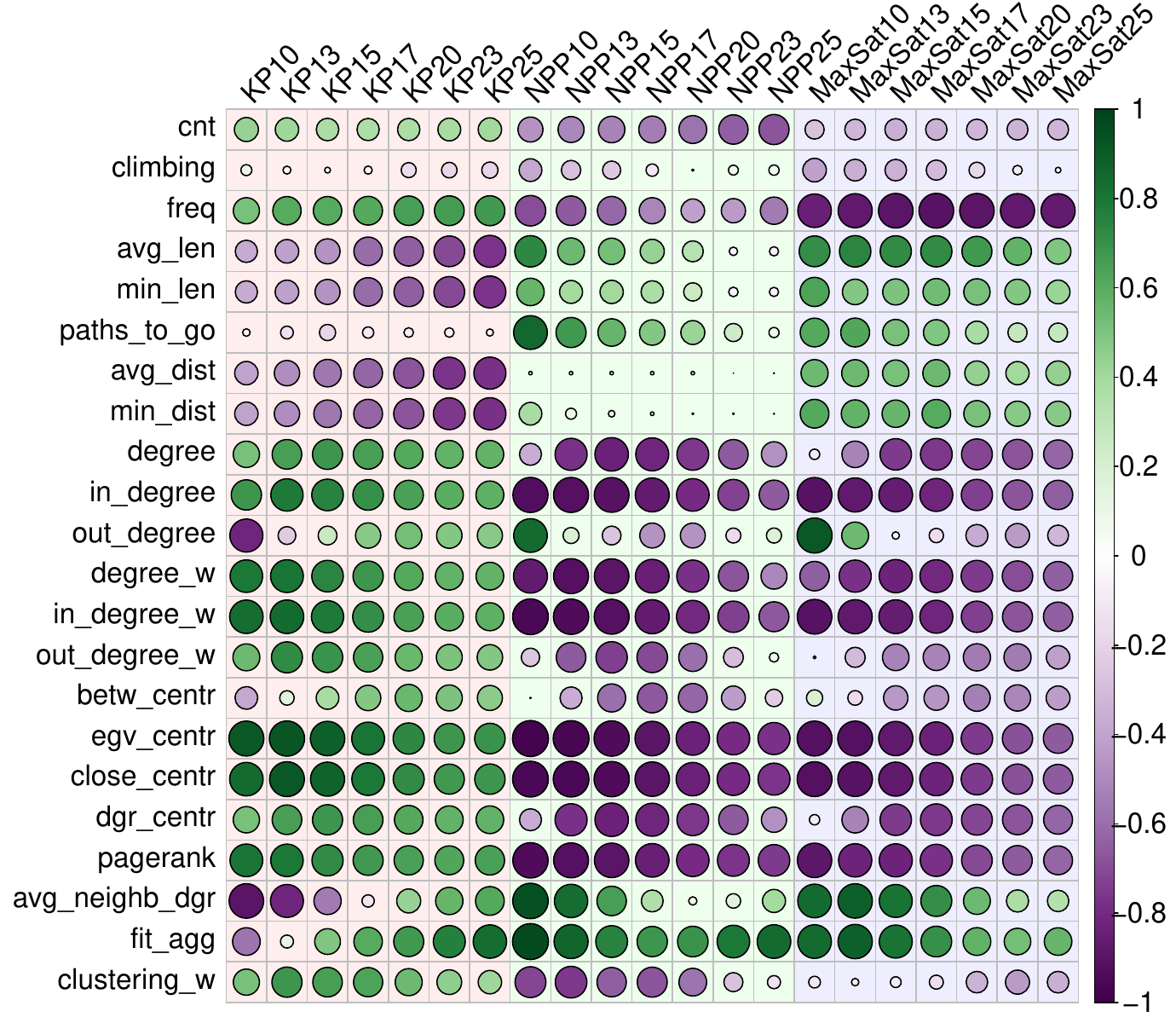}
    \caption{xxx}
    \label{fig:corr_feature}
\end{figure}

From the heatmap shown in~\pref{fig:corr_feature}, it is interesting to see that most features considered in our experiments show high correlations with the corresponding fitness values when $n\leq 23$. The only two exceptions are the average neighbor fitness/degree, which showed little relation with fitness when $n=10$. We attributed this to the high density of LON for NPP-$10$ shown in~\pref{fig:metrics}(b). This suggests each local optimum in its LON may connect to a considerable fraction of other nodes in the whole network, making it difficult to discover informative patterns through neighborhood statistics. On the other hand, as marked by the black rectangle in~\pref{fig:corr_feature}, we find that the previously identified high correlations on some selected features are diminishing when $n>23$. However, the correlations still remain strong on the other features. These observations indicate some properties are actually shared across NPP with a wide range (even all) of dimensions. Specifically, we interpret the results as follows.

\begin{itemize}
    \item All four local optima features, as shown in the first four rows of~\pref{fig:corr_feature},  exhibited a remarkable degree of negative correlation with fitness across all studied dimensions. This indicates that in all the corresponding landscapes, local optima with better fitness values require more efforts to be reached/escaped from. Also, better local optima tend to have larger basins of attraction as well as higher chances to be visited during the local search.
    \item Three features derived from LONs, as shown in the last three rows of~\pref{fig:corr_feature},  including the average neighbor fitness and the average/minimum length to global optimum, show a high positive correlation with fitness values across all dimensions. This observation suggests that local optima with better fitness values tend to be closer to the global optimum while their neighbors are likely to be solutions who are also high-quality local optima.
    \item On the other hand, other network features, as marked in the black rectangle, are highly correlated with fitness values when $n\leq{23}$. However, the relevant correlations are diminishing significantly when $n>23$. This can be attributed to the sparse distribution of the sampled local optima which we have discussed in the previous subsection, since all these indicators are highly dependent on the connectivity pattern of networks. Nevertheless, we still believe that the characteristics discovered in lower dimensions are generalizable to high dimensions in case more ILS iterations can be conducted.
\end{itemize}

\begin{tcolorbox}[sharpish corners, top=2pt, bottom=2pt, left=4pt, right=4pt, boxrule=0.0pt, colback=black!5!white,leftrule=0.75mm,]
    \textbf{\underline{Response to RQ1:}} \textit{Since the features studied in this subsection are able to capture various structural characteristics of LONs/fitness landscapes, their similar values/trajectories between problems with neighboring dimensions as well as the sharing/consistency of their correlation with local optima fitness across different dimensions, can be interpreted as structural similarity with respect to the corresponding landscapes.}
\end{tcolorbox}

\begin{figure*}[t]
    \centering
    \includegraphics[width=\linewidth]{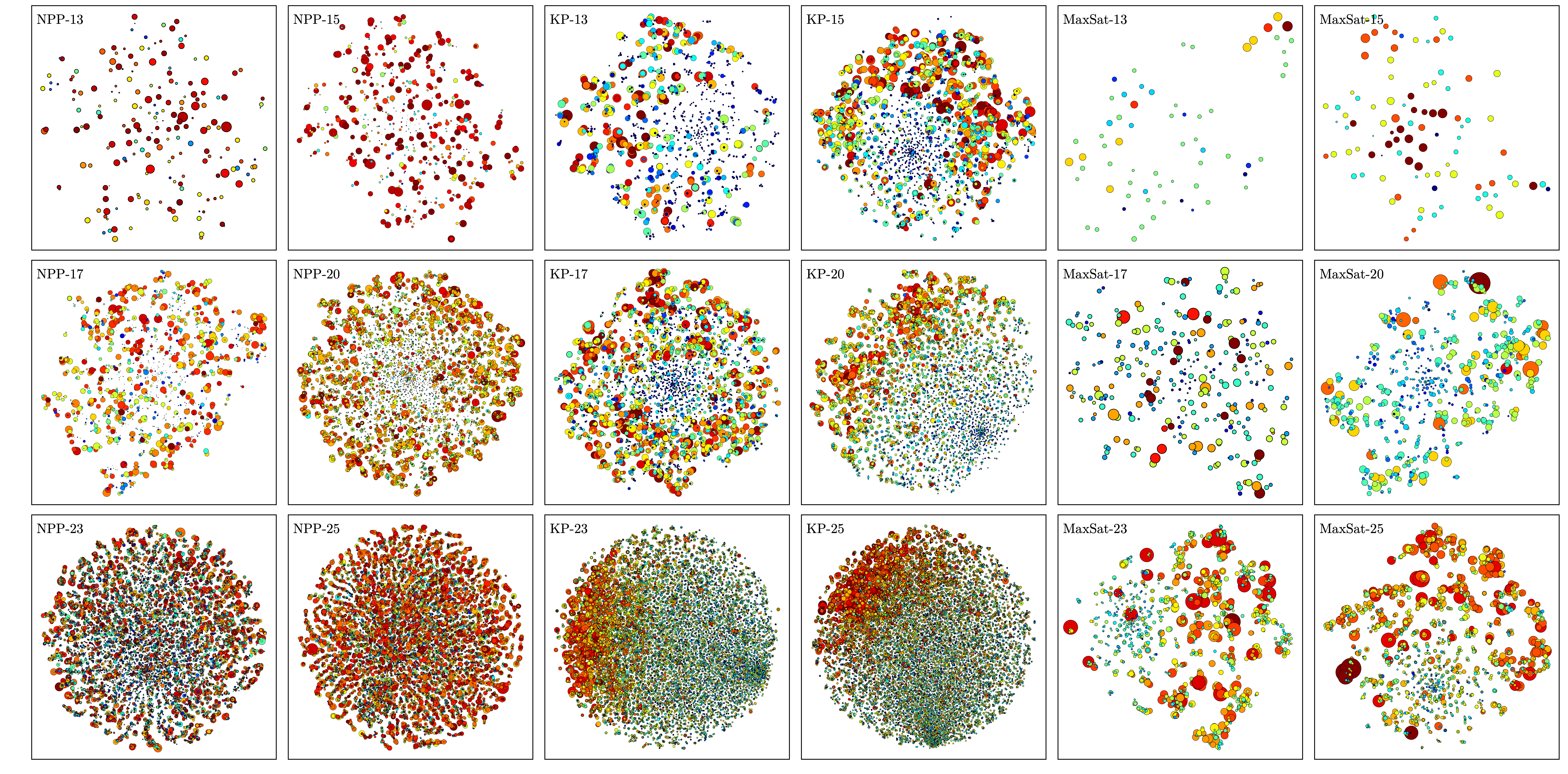}
    \caption{xxx}
    \label{fig:node_embedding}
\end{figure*}

\subsection{Structural Similarity Study for a Given Problem by Statistical Analysis}
\label{sec:rq2}

\begin{figure}[t!]
    \centering
    \includegraphics[width=.6\linewidth]{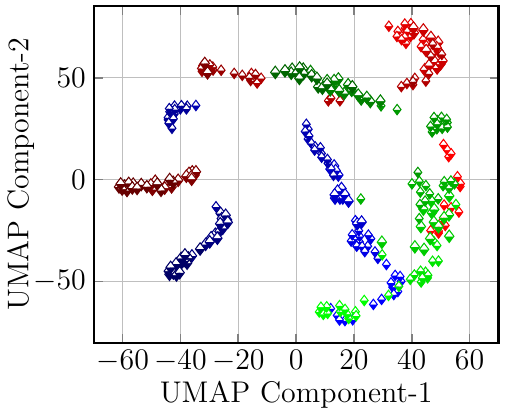}
    \caption{xxx}
    \label{fig:instance}
\end{figure}

\subsubsection{Results for visualizing topological structures of LONs}

The $2$-dimensional projections of the global structure of LONs for the three problems at different dimensions are shown in~\pref{fig:node_embedding}, from which we could observe apparent patterns across instances of a given problem with different $n$, whereas for instances belong to different kinds of problems, there is considerable difference in terms of the distribution of local optima. In particular, 

For NPP, we could see that nodes with larger degree tend to have better fitness values, which is in accordance with the results found in~\pref{fig:corr_feature}. These high-quality local optima tend to locate at the outer region of each graph, whereas as we move towards the inner regions, local optima fitness (as well as node decrease) tend to decrease. This observation generally conforms with the positive degree/fitness assortativity value obvserved for NPP with $n \geq 15$ reported in~\pref{fig:metrics} (d). 

As for KP, when the dimension is relatively low (e.g., $n \leq 15$), nodes with different fitness levels could mix together, though a central cluster consists of low-quality local optima could be observed. Recall from~\pref{fig:metrics} (d) that disassortative mixing of degree and fitness are observed for these instances. When $n$ further increases, we are able to observe a clear trend in the distribution of local optima, where ones with high-degree and high-fitness are located at the same region, and those less-fit ones form a cluster a the other side of the graph. This is also consistent with the positive assortativity value illustrated in~\pref{fig:metrics} (d).

Finally, for MaxSat, we first notice that the number of nodes in LONs are considerably fewer than the other two kinds of problems under the same dimension.  \dots.

\subsubsection{Results for visualizing LONs in bivariate instance space}

\pref{fig:instance} shows the distribution of problem instances of each problem with different dimensions in 2D instance space, where $30$ random instances are drawn for a given problem with a specific dimension.

\begin{tcolorbox}[sharpish corners, top=2pt, bottom=2pt, left=4pt, right=4pt, boxrule=0.0pt, colback=black!5!white, leftrule=0.75mm,]
    \textbf{\underline{Response to RQ2:}} \textit{By visualizing the structures of LONs for NPP at different dimensions in a latent space, we find certain regularities in terms of the distribution of local optima and their associated fitness and degree. In particular, we find similar patterns for problems in neighboring dimensions. Such observations are also reflected from an instance space analysis where LONs of instances with similar dimension(s) are close to/grouped with each other. Such observations imply the existence of structural similarity among their corresponding fitness landscapes.}
\end{tcolorbox}

\begin{figure}[t]
    \centering
    \includegraphics[width=\linewidth]{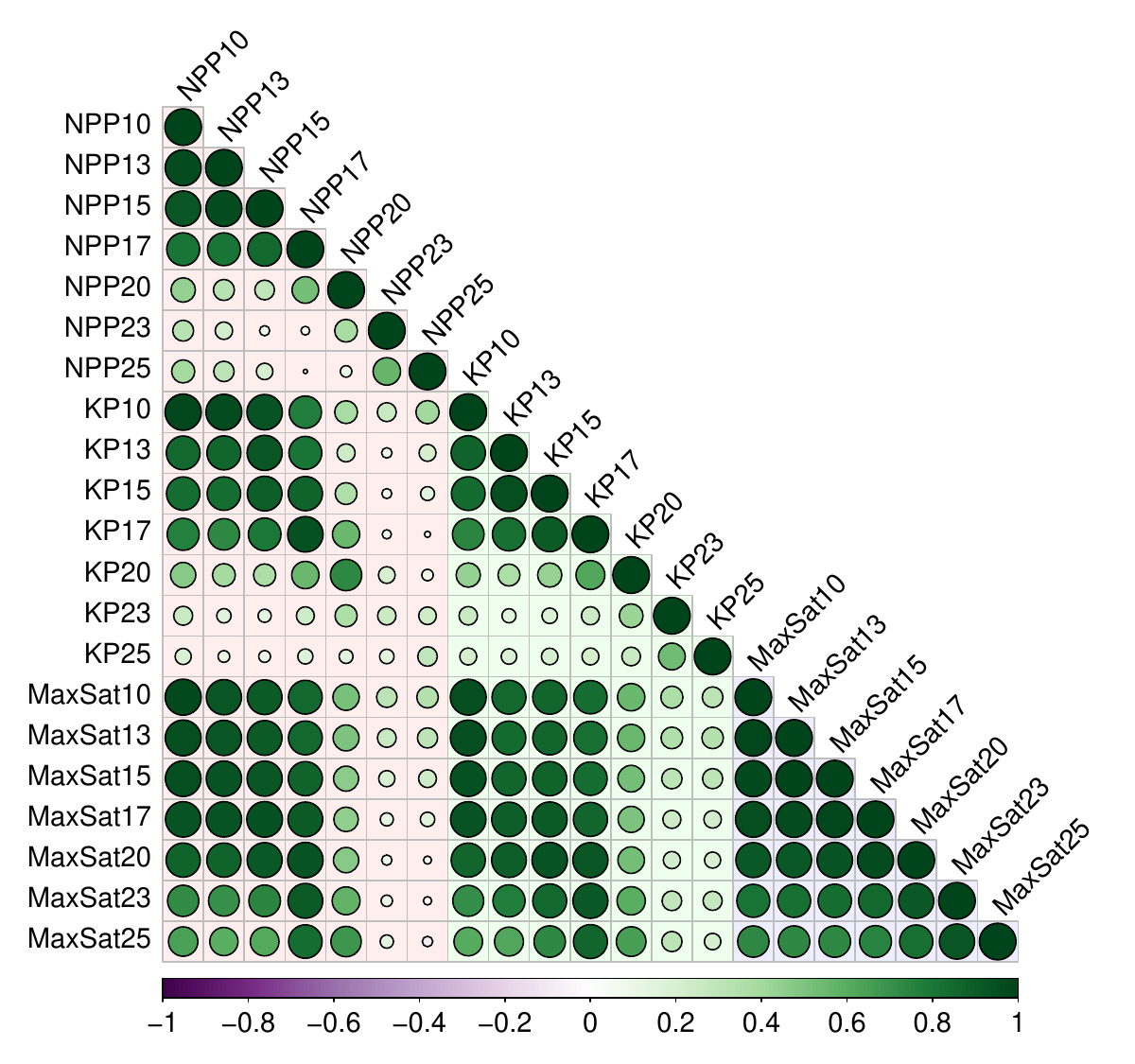}
    \caption{xxx}
    \label{fig:corr_graph}
\end{figure}

\subsection{Quantitative Measurement of Structural Similarity}
\label{sec:rq3}

The $\mathrm{Sim}$ values, i.e., the $\mathrm{Spearman}$ correlations between different problem instances, are calculated and presented as a $21\times{21}$ symmetric heatmap matrix shown in~\pref{fig:corr_graph}. These results show some level of consensus with our previous findings when responding to the first two RQs. More specifically, we find positive correlations across all dimensions. More specifically,

For any given problem, we could see that instances with a certain dimension tends to have higher correlation with instances of similar sizes, whereas they are much less correlated with those of very distinct dimensions. This is particularly obvious for NPP and KP, where we notice that low dimensional problems (e.g., $n \leq 17$) are highly similar to each other, and they have much less similarity with problems of larger sizes. As for MaxSat, generally instances with all studied dimensions are very similar to each other, yet we could still observe that instances with closer dimensions would share higher similarity.

We then compare the results across different types of problems. For

\begin{tcolorbox}[sharpish corners, top=2pt, bottom=2pt, left=4pt, right=4pt, boxrule=0.0pt, colback=black!5!white,leftrule=0.75mm,]
    \textbf{\underline{Response to RQ3:}} \textit{Our empirical study in this subsection demonstrates that graph embedding method is able to play as a driver to quantitatively evaluate the structural similarity with respect to different fitness landscapes. The observations in this subsection achieve a consensus with those discussed in~\pref{sec:rq1} and~\pref{sec:rq2}. This quantitatively consolidates the existence of structural similarity among NPPs across different dimensions.}
\end{tcolorbox}

\subsection{Effectiveness Verification of the Measured Similarity}

\subsubsection{Research methods}
\label{sec:method_rq4}

To answer RQ4, we choose simulated annealing (SA)~\cite{SA} as the meta-heuristic and test its performance on NPP instances with $n\in\{10,\cdots,20\}$. Following the ideas in~\cite{Performance_Gabriela,Performance_TEVC,Performance_K_S_Smiles}, we define the performance of SA on a certain NPP instance as the capability to find the global optimum under a fixed budget. Specifically, we run a SA on each instance for $1,000$ independent runs, and record the fraction of runs that successfully find the global optimum as the success rate ($SR$). In practice, the stopping criterion of SA is set to a fixed budget of $1,000$ iterations. In each iteration, the temperature is reduced to $T_0\times{0.8^{i/300}}$, where the initial temperature $T_0$ is set as $1,000$. We propose two metrics to measure the relative performance of SA between each pair of problem instances. One is called the difference of success rates $\Delta_\mathrm{SR}=|SR_i-SR_j|$; the other is called the relative ratio of success rates $\rho_\mathrm{SR}=SR_i/SR_j$, where $i,j\in\{10,\cdots,20\}$ and $i\neq j$. In this subsection, we first investigate the correlation between $\mathrm{Sim}$ calculated using the method in~\pref{sec:rq3} against $\Delta_\mathrm{SR}$ and $\rho_\mathrm{SR}$. Then, we apply a quadratic regression to evaluate to which extent can our calculated $\mathrm{Sim}$ explain the relative performance of SA in terms of $\Delta_\mathrm{SR}$ and $\rho_\mathrm{SR}$.

\subsubsection{Empirical results and analysis}
\label{sec:result_rq4}

\begin{figure}[t!]
    \centering
    \includegraphics[width=\linewidth]{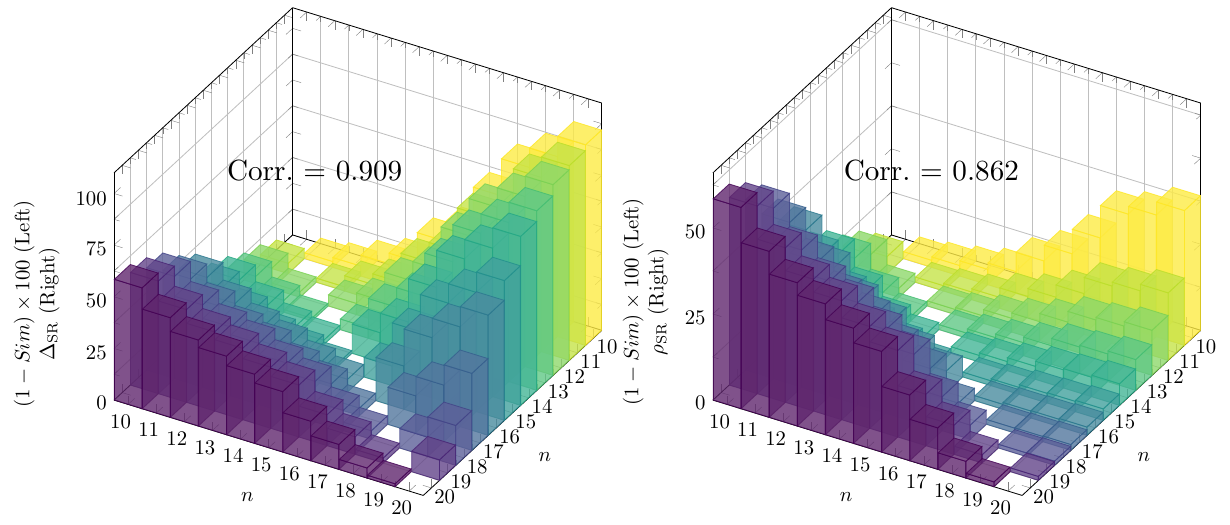}
    \caption{3D bar charts of the $\mathrm{Spearman}$ correlation coefficients of $\mathrm{Sim}$ versus $\Delta_\mathrm{SR}$ and $\rho_\mathrm{SR}$ across different dimensions ranging from $10$ to $20$. The data are averaged over $30$ random instances.}
    \label{fig:3d_bar}
\end{figure}

\begin{figure}[t!]
    \centering
    \includegraphics[width=\linewidth]{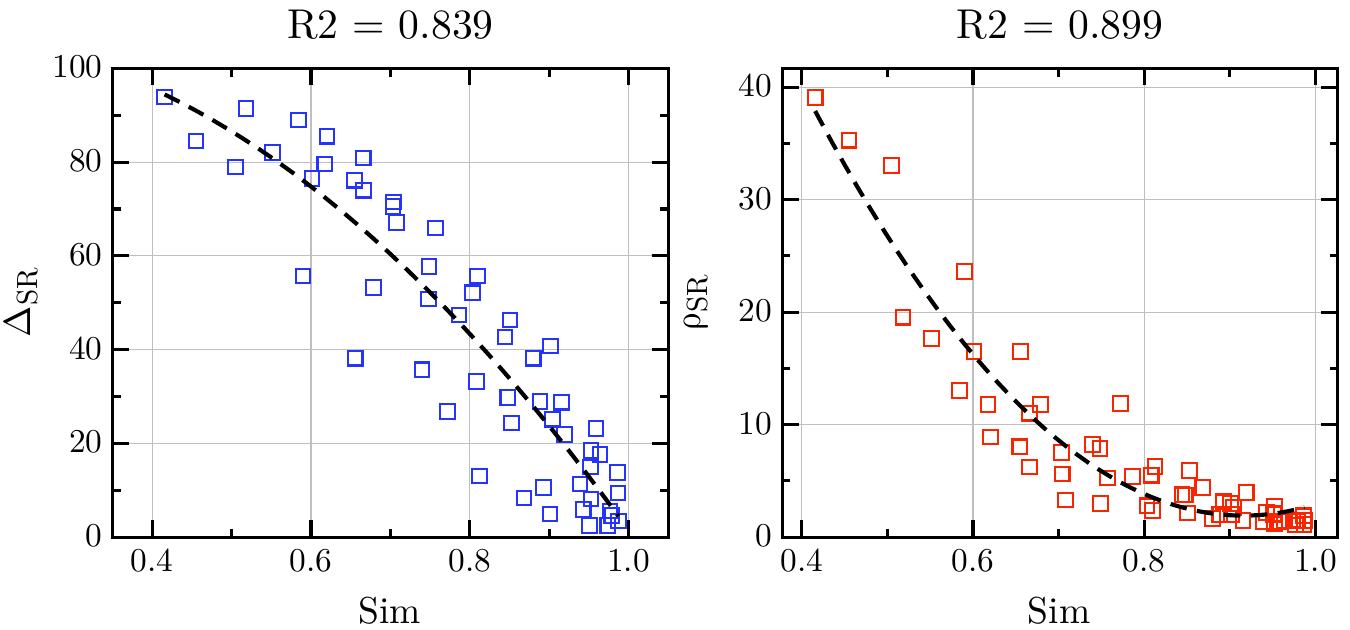}
    \caption{Quadratic regression analysis of $\Delta_\mathrm{SR}$ and $\rho_\mathrm{SR}$ versus $\mathrm{Sim}$. Each data point represents average over $30$ instances.}
    \label{fig:regression}
\end{figure}

From the bar charts shown in~\pref{fig:3d_bar}, it is clear to see both $\Delta_\mathrm{SR}$ and $\rho_\mathrm{SR}$ change in correspondence with $\mathrm{Sim}$. Their $\mathrm{Spearman}$ coefficients are $-0.909$ and $-0.862$, respectively\footnote{We transform the $\mathrm{Sim}$ to reverse its trend thus the results are shown in positive correlation values in~\pref{fig:3d_bar}.}. Furthermore, it is interesting to see the correlation between neighboring dimensions is high and it diminishes when the dimensionality becomes disparate. Such relationship is further exploited in~\pref{fig:regression} by using a quadratic regression between $\Delta_\mathrm{SR}$ and $\rho_\mathrm{SR}$ against $\mathrm{Sim}$, respectively. It is very encouraging to see both quadratic regression analyses return a high $R^2$ score close to $0.9$.

\begin{tcolorbox}[sharpish corners, top=2pt, bottom=2pt, left=4pt, right=4pt, boxrule=0.0pt, colback=black!5!white,leftrule=0.75mm,]
    \textbf{\underline{Response to RQ4:}} \textit{Here we empirically investigate the correlation between the structural similarity of fitness landscapes versus the performance of a meta-heuristic algorithm, SA in particular in our experiments. In a nutshell, SA requires similar computational cost to solve structurally similar NPPs.}
\end{tcolorbox}


\vspace{-0.9em}
\section{Conclusions and Future Directions}
\label{sec:conclusions}

By using LON as the proxy of fitness landscapes of NPP instances, this paper proposed to leverage graph data mining techniques to conduct qualitative and quantitative analyses to explore the latent topological structural information embedded in those landscapes. Our empirical results are inspiring to support the overall assumption of the existence of structural similarity between landscapes within neighboring dimensions. Besides, experiments on $\mathrm{SA}$ demonstrate that the performance of a meta-heuristic solver is similar on structurally similar landscapes.

To the best of our knowledge, this is the first attempt towards the investigation of structural similarity within combinatorial fitness landscapes. We believe graph data mining is a promising vehicle to facilitate the exploratory landscape analysis. Our methodologies are applicable for the purpose of exploring fitness landscapes of other combinatorial optimization problems. It is intriguing to see whether the structural similarity and generalizable to a wider range of combinatorial optimization problems. In addition, it is interesting to investigate the potential of the features learned and extracted by our methods for a wider range of tasks such as automatic algorithm selection and algorithm performance prediction.

\section*{Acknowledgment}
This work was supported by UKRI Future Leaders Fellowship (MR/S017062/1) and Amazon Research Awards.

\bibliographystyle{IEEEtran}
\bibliography{IEEEabrv,lon}

\end{document}